\pdfoutput=1

\documentclass[11pt]{article}

\usepackage{EACL2023}

\usepackage{times}
\usepackage{latexsym}

\usepackage[T1]{fontenc}

\usepackage[utf8]{inputenc}

\usepackage{microtype}

\usepackage{inconsolata}
\usepackage{amsmath}
\usepackage{amsfonts}
\usepackage{amssymb}
\usepackage{graphicx}
\usepackage{multirow}
\usepackage{booktabs}
\usepackage{diagbox}
\usepackage{bbding}
\usepackage{enumitem}
\usepackage{commath}
\usepackage{algorithm}
\usepackage{algorithmic}
\usepackage[toc,page]{appendix}

\makeatletter
  \newcommand\smallernormal{\@setfontsize\smallernormal{9.6pt}{9.6}}
\makeatother
\makeatletter
  \newcommand\smallernormalAppendix{\@setfontsize\smallernormalAppendix{11pt}{10}}
\makeatother

\newcommand{\bm}{\mathbf}

\DeclareMathOperator*{\argmax}{arg\,max}

 \DeclareSymbolFont{AMSb}{U}{msb}{m}{n} \DeclareSymbolFontAlphabet{\Bbb}{AMSb}
\let\mathbb\Bbb

\title{Event Temporal Relation Extraction with Bayesian Translational Model}

\author{Xingwei Tan$^1$, Gabriele Pergola$^1$, Yulan He$^{1,2,3}$ \\
  $^1$Department of Computer Science, University of Warwick, UK\\
  $^2$Department of Informatics, King's College London, UK\\
  $^3$The Alan Turing Institute, UK\\
  \texttt{\{Xingwei.Tan, Gabriele.Pergola.1\}@warwick.ac.uk}\\
  \texttt{yulan.he@kcl.ac.uk}\\}

\begin{document}
\maketitle
\begin{abstract}
Existing models to extract temporal relations between events lack a principled method to incorporate external knowledge.
In this study, we introduce \textit{Bayesian-Trans}, a Bayesian learning-based method that models the temporal relation representations as latent variables and infers their values via Bayesian inference and \textit{translational functions}. 
Compared to conventional neural approaches, instead of performing point estimation to find the best set parameters, the proposed model infers the parameters' posterior distribution directly, enhancing the model's capability to encode and express uncertainty about the predictions. Experimental results on the three widely used datasets show that Bayesian-Trans outperforms existing approaches for event temporal relation extraction. We additionally present detailed analyses on uncertainty quantification, comparison of priors, and ablation studies, illustrating the benefits of the proposed approach.\footnote{Experimental source code is available at \url{https://github.com/Xingwei-Warwick/Bayesian-Trans}}
\end{abstract}

\section{Introduction}
Understanding events and how they evolve in time has been shown beneficial for natural language understanding (NLU) and for a growing number of related tasks \cite{cheng13, Wang18summ, ning20torque,geva-etal-2021-aristotle,sun22phee}.
Howeover, events often form complex structures with each other through various temporal relations, which is challenging to track even for humans \cite{wang2020joint}.

\begin{figure}[t!]
    \centering
    \includegraphics[width=\columnwidth]{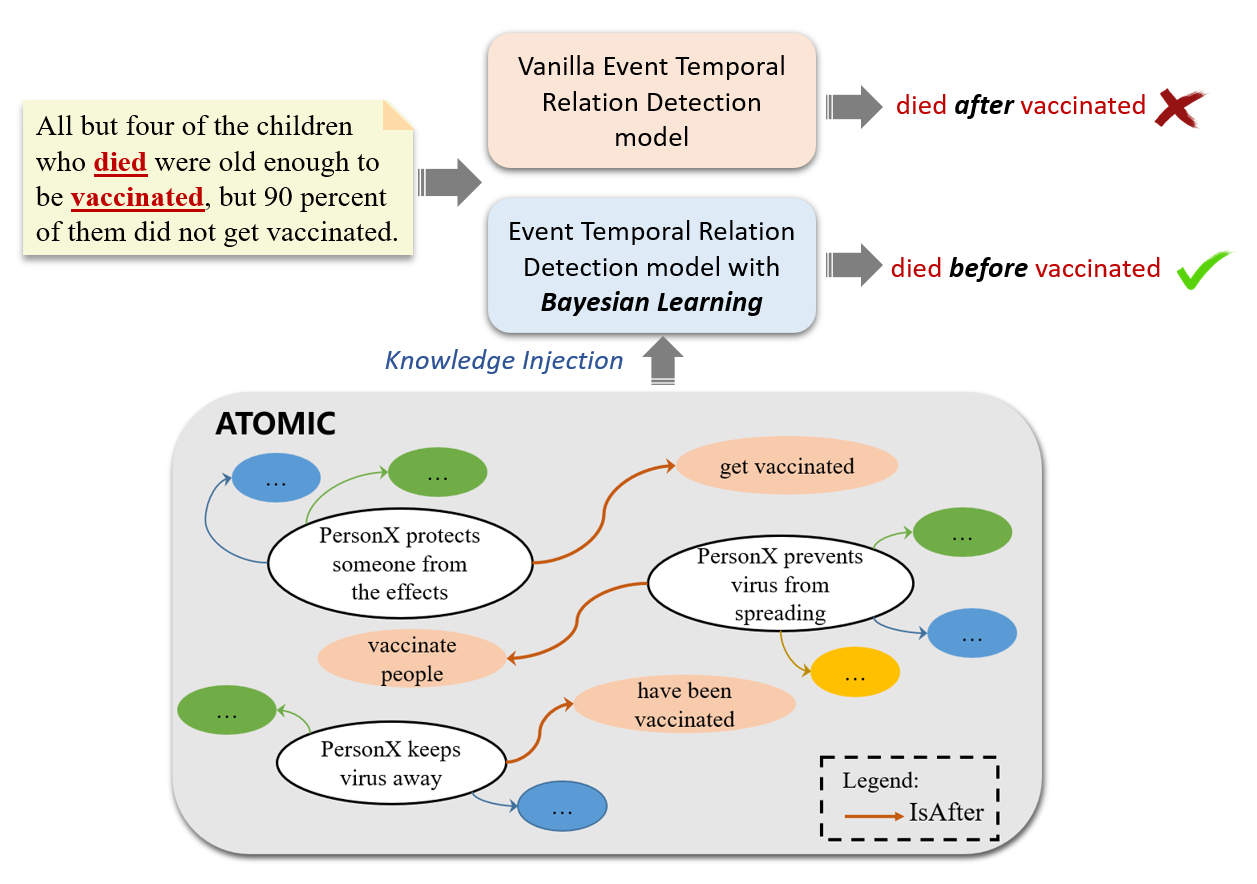}
    \caption{Comparison between with or without external knowledge incorporation on event relation extraction.}
    \label{fig:example}
\end{figure}

One of the main difficulties is the wide variety of linguistic expressions of temporal relations across different contexts. Although many of them share some linguistic similarities, most of the topics in which they occur are characterized by some shared but unspoken knowledge that determines how temporal information is expressed. 
For example, when it comes to health, prevention is widely practised, with many treatments (e.g., vaccinations) being effective only if administered \textit{before} the onset of a disorder. On the contrary, in the automotive industry, it is common that most people repair their car \textit{after} a problem occurs.  
However, despite its simplicity, such commonsense knowledge is rarely stated explicitly in text and varies greatly across different domains. 
For example, in Figure \ref{fig:example}, a 
detection model lacking the commonsense knowledge that \emph{vaccination} can protect people from infection, tends to get confused by the complex linguistic structures in the excerpt and returns the wrong prediction entailing that `\emph{died}' happens \texttt{after} `\emph{vaccinated}'. Instead, with the consideration of prior temporal knowledge involving the \emph{vaccination} event from an external knowledge source ATOMIC \cite{Hwang2021COMETATOMIC2O}, a model gives the correct prediction that `\emph{died}' occurs \texttt{before} `\emph{vaccinated}'. 

Methods proposed in recent studies for event relation extraction are mostly end-to-end neural architectures making rather limited use of such commonsense knowledge \cite{han-etal-2019-deep,han-etal-2019-joint}.
Only a few works have explored the incorporation of external knowledge to mitigate the scarcity of event annotations \cite{ning-etal-2019-improved,wang-etal-2020-joint}. 
Nevertheless, these approaches typically update the event representations with knowledge features derived from external sources, lacking a principled way of updating models' beliefs in seeing more data in the domains of interests.

In this work, we posit that the Bayesian learning framework combined with translational models can provide a principled methodology to incorporate knowledge and mitigate the lack of annotated data for event temporal relations.
Translational models, such as TransE \cite{Bordes2013TranslatingEF}, are energy-based models based on the intuition that
the relations between entities can be naturally represented by geometric translations in the embedding space.
More concretely, a relation between a \textit{head entity} and a \textit{tail entity} holds if there exists a \textit{translational} operation bringing the \textit{head} close to the \textit{tail} vector.

Specifically, we introduce a novel Bayesian Translational model (Bayesian-Trans) for event temporal relation extraction. Compared to conventional neural translational models, which only yield a point estimation of the network parameters, the Bayesian architecture can be seen as an ensemble of an infinite number of neural predictors, drawing samples from the posterior distribution of the translational parameters, refining its belief over the initial prior.
As a result, event temporal relations are determined by the stochastic translational parameters drawn from posterior distributions. Additionally, such posteriors are conditioned upon the prior learned on external knowledge graphs, providing the commonsense knowledge required to interpret more accurately the temporal information across different contexts.
As shown in the results obtained from the experimental evaluation on three commonly used datasets for event temporal relation extraction, the combination of translational models and Bayesian learning is particularly beneficial when tailored to the detection of event relations. 
Moreover, a favorable by-product of our Bayesian-Trans model is the inherent capability to express degrees of uncertainty, avoiding the overconfident predictions on out-of-distribution context.
Our contributions are summarized in the following:
\begin{itemize}
    \item We formulate a novel Bayesian translational model for the extraction of event temporal relations, in which event temporal relations are modeled through the stochastic translational parameters, considered as latent variables in Bayesian inference. 
    \item We devise and explore $3$ different priors under Bayesian framework to study how to effectively incorporate knowledge about events. 
    \item We conduct thorough experimental evaluations on three benchmarking event temporal datasets and show that Bayesian-Trans achieves state-of-the-art performance on all of them.
    We also provide comprehensive analyses of multiple aspects of the proposed model.

\end{itemize}

\section{Related Work}

This work is related to at least three lines of research: event temporal relation detection, prior knowledge incorporation, and graph embedding.

\subsection{Event Temporal Relation}
Similar to entity-level relation extraction \cite{zeng-etal-2014-relation,peng-etal-2017-cross}, the latest event temporal relation extraction models are based on neural networks, but in order to learn from limited labeled data and capture complex event hierarchies, a wide range of optimization or regularization approaches have been explored.
\citet{ning-etal-2019-improved} proposed an LSTM-based network and ensured global consistency of all the event relations in the documents by integer linear programming. 
\citet{wang-etal-2020-joint} employed RoBERTa \cite{https://doi.org/10.48550/arxiv.1907.11692} and converted a set of predefined logic rules into differentiable objective functions to regularize the consistency of the relations inferred and explore multi-task joint training.
\citet{tan-etal-2021-extracting} proposed using hyperbolic-based methods to encode temporal information in a hyperbolic space, which has been shown to capture and model asymmetric temporal relations better than their Euclidean counterparts.
\citet{hwang22} adopted instead a probabilistic box embeddings to extract asymmetric relations.
\citet{wen-ji-2021-utilizing} proposed to add an auxiliary task for relative time prediction of events described over an event timeline.
\citet{Cao2021UncertaintyAwareSF} developed a semi-supervised approach via an
uncertainty-aware self-training framework, composing a training set of samples with actual and pseudo labels depending on the estimated uncertainty scores. None of the aforementioned approaches explored Bayesian learning for incorporating prior event temporal knowledge.

\subsection{Incorporation of Prior Knowledge}
Knowledge plays a key role in understanding event relations because people often skip inessential details and express event relations implicitly which is difficult to understand without relevant knowledge.
For example, \textsc{TemProb} \cite{ning-etal-2018-improving} contains temporal relation probabilistic knowledge which is encoded by Siamese network and incorporated into neural models as additional features \cite{ning-etal-2019-improved,wang-etal-2020-joint,tan-etal-2021-extracting}.
Unlike previous works, we combine the Bayesian Neural Network with distance-based models, treating the translational parameters as latent variables to be inferred. To this end, we adopt the variational inference \cite{Kingma2014AutoEncodingVB, Blei2016VariationalIA, gui19, disen21, zhu22dis}, and derive the prior distribution of the temporal relation information from commonsense knowledge bases \cite{boost21, lu22event}.
\citet{christopoulou-etal-2021-distantly} explored a similar intuition of using knowledge base priors as distant supervision signals, but the approach and the task are different.

\subsection{Graph Embedding Learning}
Multi-relational data are commonly interpreted in terms of directed graphs with nodes and edges representing entities and their relations, respectively. Several works have recently focused on modelling these multi-relational data with relational embeddings by detecting and encoding local and global connectivity patterns between entities.

TransE \cite{Bordes2013TranslatingEF} has been a seminal work adopting geometric translations of entities to represent relations in the embedding space. If a relation between a head and a tail entity holds, it is encoded via the translational parameters learned at training time. 

However, TransE cannot model symmetry relation well by simple addition which led to several subsequent studies exploring diverse types of transformation resulting in a family of \textit{translational models} \cite{Wang2014KnowledgeGE,ji-etal-2015-knowledge,Lin2015LearningEA}.
Among them, \citet{Balazevic2019MultirelationalPG} proposed to utilize the Poincar{\'e} model, mapping the entity embeddings onto a Poincar{\'e} ball, and using the Poincar{\'e} metric to compute the score function and predict their relations. 
\citet{chami-etal-2020-low} further expanded the idea of embedding learning over manifolds by additionally considering reflections and rotations and redefining the translation over a learned manifold.

Although translational models are shown efficient in modeling graph relation, they provide relatively limited interaction between nodes than neural network-based methods, such as Graph Neural Networks \cite{estrach2014spectral,chami2020machine}.
Under this framework, nodes in a graph are neural units, which can iteratively propagate information through edges, and whose representations are learnt during the training process.
In particular, Relational Graph Convolutional Networks (RGCN) \cite{schlichtkrull2018modeling} encode relational data through link prediction and entity classification tasks, while enforcing sparsity via a parameter-sharing technique.
Although modeling knowledge graphs has been one of the main focuses of the above-mentioned graph learning approaches, they lack any systematic mechanism to inject prior knowledge and update it during training.

\begin{figure*}[tb]
    \centering
    \includegraphics[width=0.95\textwidth]{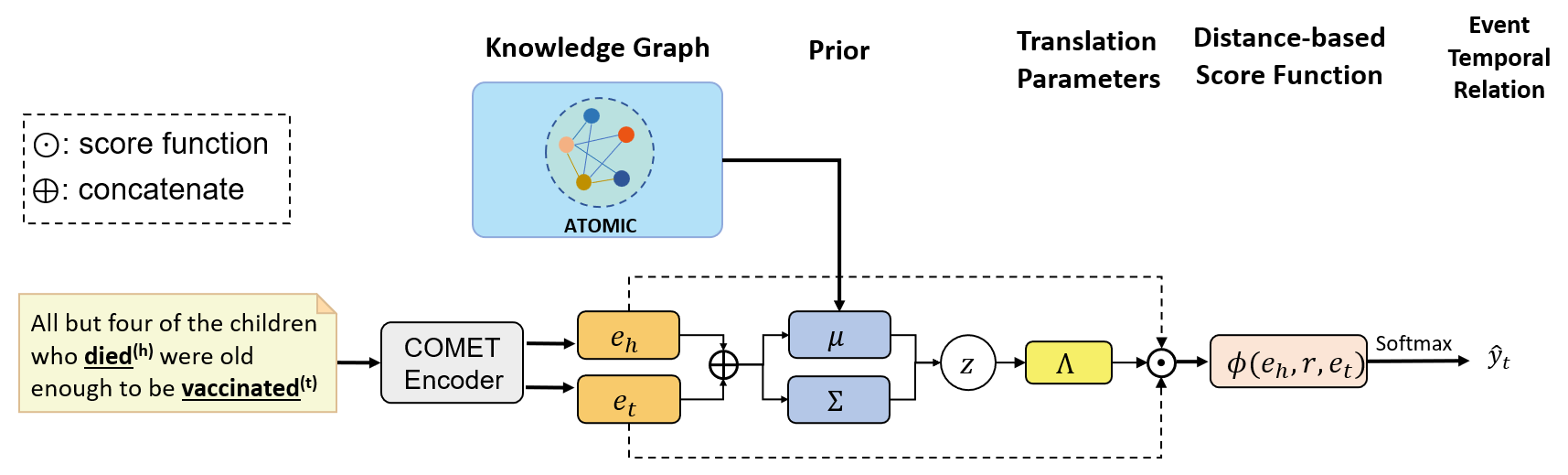}
    \caption{The network structure of Bayesian-Trans. Context sentences are first fed into a COMET encoder to generate event representations. With MLP layers, the event representations are mapped to generate a variational distribution of relation representations which is guided by KG priors. The relation representations are then used in the translational model to generate prediction scores.}
    \label{fig:model}
\end{figure*}

\section{Bayesian-Trans Model}

In identifying temporal relations between events, we aim at predicting the relation type of two events given in text, commonly denoted as \textit{head} event $x_h$ and \textit{tail} event $x_t$:
\begin{equation}
    \hat{y} = \argmax_{y\in \mathcal{R}} p(y | x_h, x_t)
\end{equation}
where $\mathcal{R}$ denotes a set of possible relation types, while $x_h$ and $x_t$ the head and tail event triggers, respectively.
Assuming that a set of latent variables $\bm{\Lambda}$ denotes the collection of all relation-specific transformation parameters $\bm{\Lambda}_r$.
For example, in the knowledge embedding learning model such as MuRE \cite{Balazevic2019MultirelationalPG}, the head entity is first transformed through a relation-specific matrix $\bm{W}_r$, followed by a relation-specific translation vector $t_r$, then $\bm{\Lambda}_r=\{\bm{W}_r, t_r\}$. By Bayesian learning, the probability of inferring a relation type $r$ can be written as:
\begin{equation}\smallernormal
    p(y=r | x_h, x_t)=\int_{\bm{\Lambda}} p(y_r | x_h, x_t, \bm{\Lambda}) p(\bm{\Lambda} | \mathcal{G}) d_{\bm{\Lambda}}\label{eq:relationPred}
\end{equation}
Here, $p(\bm{\Lambda} | \mathcal{G})$ denotes the prior distribution of $\bm{\Lambda}$ derived from an existing knowledge graph encoded as $\mathcal{G}$. Directly inferring Eq. (\ref{eq:relationPred}) is intractable. But we can resort to amortised variational inference to learn model parameters. In what follows, we present our proposed Bayesian learning framework built on translational models for event temporal relation extraction, called \textbf{Bayesian-Trans}, with its architecture shown in Figure~\ref{fig:model}.

In particular, the context $S$ in which the two events occur is the input to our Bayesian-Trans. First, we encode $S$ via a pre-trained language model generating the contextual embeddings ${\rm{e}}_h$ and ${\rm{e}}_t$ for the triggers of the head and tail events, respectively. The contextualised event trigger representations, ${\rm{e}}_h$ and ${\rm{e}}_t$, are fed as input into a Bayesian translational module. This module, by means of variational inference, determines the parameters of the translational model, encoding the posterior distribution of the temporal relations conditioned upon the input events. Finally, we use a score function on the translated head and tail triggers to predict their temporal relation. 
We provide a more detailed description in the following.

\subsection{Contextual Encoder}
The proposed model uses COMET-BART \cite{Hwang2021COMETATOMIC2O} as the context encoder. COMET-BART is a BART pre-trained language model \cite{lewis-etal-2020-bart} fine-tunned on ATOMIC \cite{bosselut-etal-2019-comet,Hwang2021COMETATOMIC2O}, which is an event-centric knowledge graph encoding inferential knowledge about entities and events, including event temporal relations.
The COMET-BART is able to generate consequence events given the antecedent event and a relation with good accuracy thus is regarded encodes knowledge well.
Following the approach adopted in previous works \cite{ning-etal-2019-improved, wang-etal-2020-joint, tan-etal-2021-extracting}, we use the representation of the first token of an event trigger as the contextual embedding of that event\footnote{We conducted some exploratory experiments adopting the last token or the average representation, but results showed that the first token was still the best option in this context.}, ${\rm{e}}_h, {\rm{e}}_t=\textrm{COMET-BART}(x_h, x_t)$, where ${\rm{e}}_h,{\rm{e}}_t\in\mathbb{R}^d$.
The event representations are then concatenated together and fed through MLPs to generate the parameters of the variational distribution, from which the latent event-pair representation $z$ is sampled.
$z$ is then mapped to the parameter space of the translational model as $\bm{\Lambda}$.

\subsection{Incorporating Knowledge via Bayesian Learning}
The proposed model utilizes relation embeddings for classifying event relation in a similar manner as the translational models in knowledge graph embedding, such as TransE \cite{Bordes2013TranslatingEF}.
If the embedding of the tail event is close enough to the embedding of head event after applying a series of relation-specific transformation, the relation stands, and vice versa.
A wide range of translational models typically proposed for learning knowledge graph embeddings can be adopted in the proposed Bayesian-Trans.
Additionally, to incorporate prior knowledge, we extend translational models to operate within the Bayesian inference framework. We proceed with introducing a standard translational model in the context of temporal relations, and describe how we extend it to work in the Bayesian framework.

\paragraph{Translational Model}
Generally speaking, a translational model uses \textit{relation representations} $\bm{\Lambda}_r$ to perform ``translation'' for relation $r$ on the head and tail events. Then, the transformed head and tail event embeddings are compared using a \textit{distance-based score function}, whose score is indicative of the temporal relation between the events. 
The score function $\phi(\cdot)$ takes the general form:
\begin{equation}
    \phi({\rm{e}}_h, r, {\rm{e}}_t) = -d(\mathcal{T}^h_{\bm{\Lambda}_r}({\rm{e}}_h),\mathcal{T}^t_{\bm{\Lambda}_r}({\rm{e}}_t))
\end{equation}
where $r$ is a relation type, $\mathcal{T}_{\bm{\Lambda}_r}(\cdot)$ is a function depending on the parameters $\bm{\Lambda}_r$ of relation $r$ to transform the event embeddings $e_h$ and $e_t$, and $d(\cdot)$ is any distance metrics (e.g., Euclidean distance). 
We explored several models with different translation functions and distance metrics in the context of temporal relations, including TransE \cite{Bordes2013TranslatingEF}, AttH \cite{chami-etal-2020-low}, MuRE \cite{Balazevic2019MultirelationalPG} and MuRP \cite{Balazevic2019MultirelationalPG}, and based on our preliminary results\footnote{Experimental results using different translational models are shown in Table \ref{table:dif_trans}.}, we eventually adopted MuRE as it strikes a good balance of training efficiency and accuracy of temporal relation classification.
We define the scoring function in the proposed model as follows:
\begin{equation}
    \phi({{\rm{e}}}_h, r, {\rm{e}}_t) = -\norm{\bm{W}_{r} {\rm{e}}_h + {\rm{t}}_{r} - {\rm{e}}_t}_2^2 
\end{equation}
where $\bm{W}_{r}\in\mathbb{R}^{d\times d}$ is a diagonal relation matrix and $t_r\in\mathbb{R}^d$ a translation vector of relation $r$, $\bm{\Lambda}_r=\{\bm{W}_{r},{\rm{t}}_{r}\},r\in \mathcal{R}$.

Although the number of parameters to train is rather low, the number of annotated samples is usually small compared to the wide range of linguistic expressions capturing temporal relations. 
We thus extend the MuRE model into a Bayesian framework to enhance its scalability by treating the translational parameters $\bm{\Lambda}$ as latent variables. The proposed framework enhances generalization by defining a variational inference process that optimizes the regularization and leverages the additional information injected via the prior distributions.

\paragraph{Bayesian Inference}
As shown in the inference equation \ref{eq:relationPred}, the prior is derived from an external knowledge graph, such as ATOMIC, as a means to inject prior information about events and temporal relations. 
In particular, $\bm{\Lambda}$ is assumed to follow a Gaussian distribution with unit variance and with mean determined by the relation representations trained on the knowledge graph. The probability function is formulated as a softmax function over a pre-defined scoring function:

{\smallernormal
\begin{equation}
    p(y_r | {\rm{e}}_h, {\rm{e}}_t, \bm{\Lambda}) = \frac{\exp\big(\phi({\rm{e}}_h, r, {\rm{e}}_t)\big)}{\sum_{r'\in\mathcal{R}}\exp\big(\phi({\rm{e}}_h, r', {\rm{e}}_t)\big)}
\end{equation}
}
\noindent with ${\rm{e}}_h$ and ${\rm{e}}_t$ denoting the embedding for the head and the tail events, respectively.

Yet, Eq. (\ref{eq:relationPred}) is intractable and cannot be inferred directly.
Thus, we resort to amortized variational inference by introducing a variational posterior $q_\theta(\bm{\Lambda} | x_h, x_t)$, which follows the isotropic Gaussian distribution and can be modeled as:

{\smallernormal
\begin{equation}
\begin{gathered}
\mu = f_{\mu}({\rm{e}}_h; {\rm{e}}_t) \quad
\Sigma = \mbox{diag}\big(f_{\Sigma}({\rm{e}}_h; {\rm{e}}_t)\big) \\
q_{\theta}(\bm{\Lambda} | {\rm{e}}_h, {\rm{e}}_t) = \mathcal{N}(\bm{\Lambda}| \mu, \Sigma),
\label{eq:variationalPosterior}
\end{gathered}
\end{equation}
}
\noindent where $f_{\mu}$ and $f_{\Sigma}$ are both fully connected layers that map the event pair representation into the parameters of the variational distribution.

Following the amortized variational inference, we maximize the evidence lower bound (ELBO) $\mathcal{L}_{e}$, defined in Eq. (\ref{eq:ELBO}), and approximated by a Monte Carlo estimation with sample size $N$, as described in Eq. (\ref{eq:MCEstimates}):

{\smallernormal
\begin{align}
\mathcal{L}_{e} &= \mathbb{E}_{q_{\theta}(\bm{\Lambda} |x_h, x_t), \{x_h, x_t\}\in\mathcal{D}} \big[\log p_{\theta}(y | x_h, x_t, \bm{\Lambda})\big] - \nonumber\\
&\quad\mbox{Reg}\big(q_{\theta}(\bm{\Lambda} | x_h, x_t, \mathcal{G} ) || p(\bm{\Lambda} | \mathcal{G} ) \big) \label{eq:ELBO} \\
&\approx \frac{1}{N}\sum_{n=1}^N\sum_{\{x_h, x_t\}\in\mathcal{D}} \big[\log p_{\theta}(y | x_h, x_t, \bm{\Lambda}^{(n)}) - \nonumber \\
&\quad\mbox{Reg}\big(q_{\theta}(\bm{\Lambda}^{(n)} | x_h, x_t, \mathcal{G} ) || p(\bm{\Lambda}^{(n)} | \mathcal{G} ) \big)  \big] \label{eq:MCEstimates}
\end{align}}
\noindent where $\mbox{Reg}(\cdot)$ is a regularization term which will be discussed in \ref{prior_and_regular}.
To train end-to-end a fully differentiable model, we adopt the reparameterization trick \cite{KingmaW14}.

\subsection{Prior Distribution and Regularization}
\label{prior_and_regular}
We proceed to discuss how the Bayesian framework enabled the incorporation of prior acquired from an external knowledge source. 
Then, we provide the details of how we compute the regularization term to induce a more stable training.

\paragraph{Prior Distribution}
One of the main advantages of the Bayesian inference framework is the possibility to inject commonsense knowledge into the model through the prior distribution of the latent variables, i.e., $p(\bm{\Lambda} | \mathcal{G})$ in Eq. (\ref{eq:relationPred}), where $\bm{\Lambda}$ are the translational parameters and $\mathcal{G}$ denotes an external knowledge graph, in our case, the ATOMIC knowledge graph \cite{Hwang2021COMETATOMIC2O}. ATOMIC is a commonsense knowledge graph containing inferential knowledge tuples about entities and events encoding social and physical aspects of human everyday experiences. For our task of event temporal relation extraction, we are only interested in the events linked via temporal relations, such as `\textsc{IsBefore}' (23,208 triples) or `\textsc{IsAfter}' (22,453 triples). By conducting link prediction on these links, we use relation embeddings learnt using an RGCN \cite{schlichtkrull2018modeling} as the mean of the prior distribution for the translational latent variables. For the relations in the experiment dataset that do not have applicable counterparts in ATOMIC (e.g., \textsc{Vague}), we set their priors to standard Gaussian. The variance of the priors is defined as the identity matrix. 

Specifically, we use COMET-BART to encode the event nodes from ATOMIC, then use their context embeddings as the node features in the RGCN.
In our preliminary experiment, we also found that RGCN cannot train well on the commonsense graph with only the event-event relation links. The graph is too sparse which makes the information difficult to propagate through the nodes. Thus, we added semantic similarity links based on the cosine similarity of the event context embeddings.
During the training of the RGCN, the node embeddings are kept frozen.
After the training of the link prediction task, we extract the relation embeddings of the RGCN.

\paragraph{Regularization Term}

To mitigate the posterior collapse problem \cite{NEURIPS2019_7e3315fe} and have a stable inference process, we adopt the Maximum Mean Discrepancy (MMD)\footnote{MMD calculation can be found in Appendix \ref{appendix:mdd}.} which is an estimation of Wasserstein distance \cite{tolstikhin2018wasserstein} as the regularization term (Eq. \ref{eq:MCEstimates}).

\section{Experimental Setup}

\paragraph{Datasets}
We evaluated the proposed Bayesian-Trans model on three event temporal relation datasets: MATRES \cite{ning-etal-2018-multi}, Temporal and Causal Reasoning (TCR) \cite{ning-etal-2018-joint}, and TimeBank-Dense (TBD) \cite{cassidy-etal-2014-annotation}. 
TimeBank-Dense is a densely annotated dataset focusing on the most salient events and providing $6$ event temporal relations. 
MATRES follows a new annotation scheme which focuses on main time axes, with the temporal relations between events determined by their endpoints, resulting in a consistent inter-annotator agreement (IAA) on the event annotations \cite{ning-etal-2018-multi}.
TCR follows the same annotation scheme, yet with a much smaller number of event relation pairs than in MATRES.
Table \ref{data_table} shows the statistics of the datasets. 

\begin{table}[htb]\small
  \begin{center}
  \resizebox{\columnwidth}{!}{
  \begin{tabular}{lrrr}
  \toprule
  \bf Class & MATRES & TCR & TBD \\ 
  \midrule 
  \textsc{Before} & $6,852$ & $1,780$ & $2,590$\\
\textsc{After} & $4,752$ & $862$ & $2,104$\\
\textsc{Equal}/\textsc{Simultaneous} & $448$ & $4$ & $215$\\
\textsc{Vague}/\textsc{None}  & $1,425$ & N/A & $5,910$\\
\textsc{Include} & N/A & N/A & $836$\\
\textsc{IsIncluded} & N/A & N/A & $1,060$\\
\midrule 
Total & $12,740$ & $2,646$ & $12,715$ \\
  \bottomrule
  \end{tabular}}
  \end{center}
  \caption{The statistics of MATRES, TCR, and TBD.}
  \label{data_table}
\end{table}

\paragraph{Baselines}
We compare the proposed Bayesian-Trans\footnote{Hyperparameter setting can be found in Appendix \ref{appendix:hyperparameters}.} with the following baselines:\\

\noindent\underline{CogCompTime} \cite{ning-etal-2018-cogcomptime} is a multi-step system which detect temporal relation using semantic features and structured inference.

\noindent\underline{BiLSTM} is a basic relation prediction model built by \citet{han-etal-2019-joint}.

\noindent\underline{LSTM + knowledge} \cite{ning-etal-2019-improved} incorporates knowledge features learnt from an external source and optimize global consistency by ILP.

\noindent\underline{Deep Structured} \cite{han-etal-2019-deep} adds a structured support vector machine on top of a BiLSTM.

\noindent\underline{Joint Constrained Learning} \cite{wang-etal-2020-joint} constrains the training of a RoBERTa-based event pair classifier using predefined logic rules, while knowledge incorporation and global optimization are also included.

\noindent\underline{Poincaré Event Embedding} \cite{tan-etal-2021-extracting} learns event embeddings based on a Poincaré ball and determines the temporal relation base on the relative position of events.

\noindent\underline{HGRU + knowledge} \cite{tan-etal-2021-extracting} is a neural architecture processing temporal relations via hyperbolic recurrent units which also incorporates knowledge features like \underline{LSTM + knowledge}.

\noindent\underline{Relative Event Time} \cite{wen-ji-2021-utilizing} is a neural network classifier combining an auxiliary task for  relative time extraction over an event timeline.  

\noindent\underline{UAST} \cite{Cao2021UncertaintyAwareSF} is an uncertainty-aware self-training model. We show the result of the model which is trained on all the labeled data.

\begin{table*}[t!]\small
  \begin{center}
  \resizebox{0.9\textwidth}{!}{
  \begin{tabular}{lcccccc}
  \toprule
  ~ & \multicolumn{3}{c}{MATRES} & \multicolumn{3}{c}{TCR} \\
  \cmidrule(lr){2-4} \cmidrule(lr){5-7}
  \bf Model & P & R 
  &  F\textsubscript{1} & P & R 
  &  F\textsubscript{1} \\ 
  
  \midrule
  CogCompTime \cite{ning-etal-2018-cogcomptime} & $61.6$ & $72.5$ & $66.6$ & - & - & $70.7$ \\
  Poincaré Event Embeddings \cite{tan-etal-2021-extracting} & $74.1$ & $84.3$ 
  & $78.9$ & $85.0$ & $86.0$ 
  & $85.5$ \\

    Relative Event Time \cite{wen-ji-2021-utilizing} & $78.4$ & $85.2$ 
    & $81.7$

    &$84.3$ & $\bf{86.8}$ 
    &  $85.5$ \\
    \midrule
  LSTM + knowledge \cite{ning-etal-2019-improved} & $71.3$ & $82.1$ 
  & $76.3$ & - & - 
  & $78.6$  \\
  Joint Constrainted Learning \cite{wang-etal-2020-joint} & $73.4$ & $85.0$ 
  & $78.8$ & $83.9$ & $83.4$ 
  & $83.7$  \\
  HGRU + knowledge \cite{tan-etal-2021-extracting} & $79.2$ & $81.7$ 
  & $80.5$ & $88.3$ & $79.0$ 
  &  $83.5$ \\
    \midrule
    Bayesian-Trans & $\bf{79.6}$ & $\bf{86.0}$ 
    & $\bf{82.7}$ & $\bf{89.8}$ & $82.6$ 
    &  $\bf{86.1}$ \\

  \bottomrule
  \end{tabular}}
  \end{center}
  \caption{Experimental results on MATRES and TCR. The first three lines contain methods without commonsense knowledge incorporation. The rest are methods which inject commonsense knowledge. The results of \citet{wang-etal-2020-joint} and \cite{wen-ji-2021-utilizing} on TCR are generated from our run of the source code provided by the authors since they are not available in their original papers.  
  The others are taken from the cited papers.}
  \label{matres_result}
\end{table*}

\begin{table}[tb]
  \begin{center}
  \resizebox{\columnwidth}{!}{
  \begin{tabular}{lc}
  \toprule
  \bf Model & Micro-F\textsubscript{1}  \\ 
  \midrule
  BiLSTM \cite{han-etal-2019-joint} & $61.9$ \\
  Deep Structured \cite{han-etal-2019-deep} & $63.2$ \\
  Relative Event Time \cite{wen-ji-2021-utilizing} & $63.2$ \\
  UAST \cite{Cao2021UncertaintyAwareSF} & $64.3$ \\
    \midrule
    Bayesian-Trans & $\bm{65.0}$ \\ 

  \bottomrule
  \end{tabular}}
  \end{center}
  \caption{Experimental results on TBD. All compared methods do not incorporate commonsense knowledge explicitly. The result of \citet{wen-ji-2021-utilizing} is generated from our run of the source code provided by the authors since they are not available in their original paper. The others are taken from the cited papers.}
  \label{tbd_result}
\end{table}

\section{Experimental Results}
\label{sec:exp}
\paragraph{Temporal Relation Classification}

We first compare Bayesian-Trans with the most recent approaches for temporal event classification in Table \ref{matres_result}, including methods with or without commonsense knowledge injection. The results are obtained by training models on the MATRES training set and evaluated on both the MATRES test set and TCR.
Table \ref{tbd_result} shows results from the TBD dataset which are generated using the provided train, development, and test sets.
We report F\textsubscript{1} score on MATRES and TCR following the definition in \cite{ning-etal-2019-improved}, and micro-F\textsubscript{1} on TimeBank-Dense.
Compared with existing methods, the proposed Bayesian-Trans has generally better performance on all three datasets, with more noticeably improvements on MATRES.
Bayesian-Trans has significant performance gains over previous methods with knowledge incorporation, which shows that it can utilize knowledge more extensively.
Details of the per-class performance can be found in Table \ref{table:class-wise} and \ref{table:class-wise_tbd}.

\paragraph{Ablation Study}
We conducted an ablation study to highlight the impact of the different modules composing Bayesian-Trans.
The results are shown in Table \ref{table:ablation}. 
In particular, we have the following variants: (1) RoBERTa$+$MLP, using RoBERTa to encode the context and then feeding representations of head and tail events to a multi-layer perceptron (MLP) for temporal relation classification; (2) RoBERTa$+$ Vanilla MuRE, using MuRE to extract temporal relations without modeling its parameters as latent variables; (3) RoBERTa$+$Bayesian-Trans, our proposed model by replacing COMERT-BART with RoBERTa as the text encoder; (4) COMET-BART$+$MLP, using COMET-BART as context encoder and an MLP for temporal relation classification; and (5) COMET-BART$+$ Vanilla MuRE, the proposed model without Bayesian learning or knowledge incorporation. 
The results demonstrate that COMET-BART is a better choice as the context encoder. Using MuRE for event temporal knowledge embedding learning does not bring any improvement compared to using a simple MLP layer for event temporal relation prediction (see (1) cf. (2), and (4) cf. (5)). 
Regardless of the contextual encoder used, the results of (3) and (6) show the benefit of employing Bayesian learning which naturally incorporates prior knowledge of event temporal relations learned from an external knowledge source for event temporal relation detection. With our proposed Bayesian translational model, we observe an improvement of $0.9-1.8\%$ in micro-F\textsubscript{1} on MATRES and $0.2-2.5\%$ in micro-F\textsubscript{1} on TimeBank-Dense compare to their non-Bayesian counterparts.

\begin{table}[tb]
  \begin{center}
  \resizebox{\columnwidth}{!}{
  \begin{tabular}{lcc}
  \toprule
  \bf Model & \bf MATRES & \bf TBD \\ 
  \midrule 
  (1) RoBERTa $+$ MLP &  $81.5$ & 62.8\\
  (2) RoBERTa $+$ Vanilla MuRE & $80.4$ & 60.5\\
  (3) RoBERTa $+$ Bayesian-Trans & $82.2$ & 63.0\\
  (4) COMET-BART $+$ MLP & $81.8$ & 63.2\\
  (5) COMET-BART $+$ Vanilla MuRE & $81.8$ & 62.6\\
  
  \midrule
  (6) COMET-BART $+$ Bayesian-Trans & $\bf{82.7}$ & $\bf{65.0}$\\
  \bottomrule
  \end{tabular}}
  \end{center}
  \caption{Ablation test results on MATRES and TBD.} 
  \label{table:ablation}
\end{table}

\begin{figure}[tb]
    \centering
    \includegraphics[width=\columnwidth]{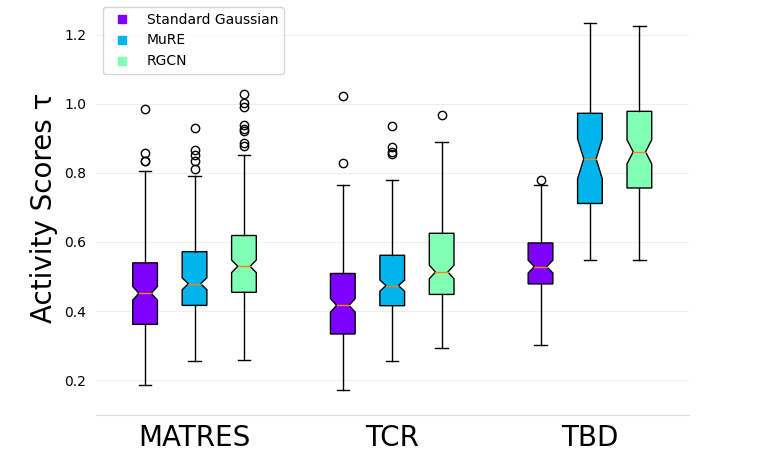}
    \caption{The box chart of the activity scores across all the dimensions of the latent encoding $\bm{\Lambda}$ with respect to the priors used in the model.}
    \label{fig:activity}
\end{figure}

\paragraph{Effects of the Priors}
\label{effect_of_priors}
We further investigate the impact of different priors on the model performance.
Inspired by the work on VAEs by \citet{burda2016importance} and \citet{truong2021bilateral}, we employed an \textit{`activity' score}, ${\tau=Cov_{e_h,e_t}(\mathbb{E}_{q(\bm{\Lambda}|e_h,e_t)}[\bm{\Lambda}])}$ to evaluate the quality and diversity of the latent encodings.
The intuition behind the ``activity'' score is that if a latent dimension encodes relevant information and is not redundant, its value is expected to vary significantly over different inputs.
By computing the score across all the test instances, every dimension of $\bm{\Lambda}$ is given an `activity' value.
Latent units with a higher value are considered more active and thus more informative.
Figure \ref{fig:activity} shows activity scores with respect to different prior distributions, including the standard Gaussian prior and priors learned on ATOMIC using MuRE or RGCN, in which the latent variables are the least active when using standard Gaussian as the prior distribution.
The higher activation is obtained using the priors learnt on the external knowledge base. In particular, the prior based on RGCN and MuRE over ATOMIC displays the most active units, with RGCN showing the most active units on average.
Table \ref{table:priors} shows the performance of the proposed model based on different priors.
Two-sided Welch's t-test ($p<0.05$) also supports that the RGCN-learned prior improves over standard Gaussian prior.

\begin{table}[tb]
  \begin{center}
  \resizebox{\columnwidth}{!}{
  \begin{tabular}{p{0.1\textwidth}>{\centering}p{0.2\textwidth}>{\centering}p{0.1\textwidth}>{\centering\arraybackslash}p{0.1\textwidth}}
  \toprule
  \bf Dataset & Standard Gaussian & MuRE & RGCN \\ 
  \midrule
  MATRES & $81.2$ & $81.8$ & $82.7$ \\
  TCR & $84.3$ & $85.4$ & $86.1$ \\
  TBD & $63.6$ & $64.6$ & $65.0$ \\
  \bottomrule
  \end{tabular}}
  \end{center}
  \caption{F\textsubscript{1} values based on different priors used in the proposed model.} 
  \label{table:priors}
\end{table}


\paragraph{Uncertainty Quantification}
\label{sec:UQ}
We present an analysis of uncertainty quantification of the Bayesian-Trans predictions.
We adopted the uncertainty quantification methods as in \citet{malinin2018predictive}, computing the entropy (\textit{total uncertainty}) and mutual information (\textit{model uncertainty}) to visualize the predictive probabilities on a 2-simplex.
Each forward pass on the same test instance is represented as a point on the simplex.
For the sake of clarity of the visualization, we removed the \textsc{Equal} class, which is hardly ever predicted by the models.

\begin{figure}[tb]
    \centering
    \includegraphics[width=\columnwidth]{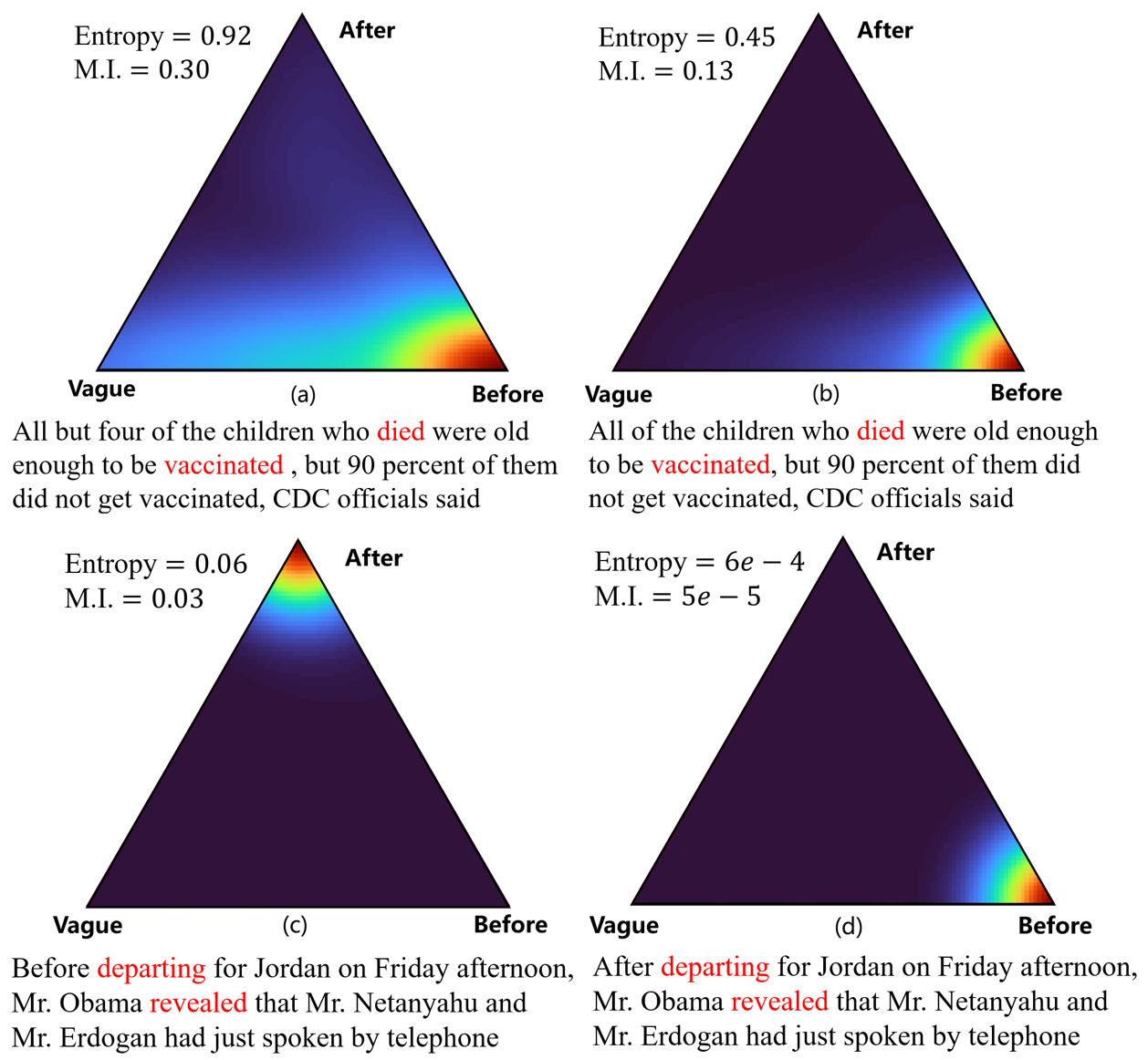}
    \caption{Examples of temporal relations in text and uncertainty quantification (entropy and mutual information) for the Bayesian-Trans model. Examples (\textit{a}),(\textit{b}) show how simplifying the linguistic structure without altering the temporal relation increases the model confidence. While examples (\textit{c}),(\textit{d}) illustrate the model's detection of temporal linguistic hints and its confidence.\vspace{-10pt}
    }
    \label{fig:uncertain}
\end{figure}

In one of the test cases (Figure~\ref{fig:uncertain}(a)), the true label is ``\emph{die}'' \textsc{before} ``\emph{vaccinate}''.
This example exhibits a rather complex linguistic structure, as such, the model exhibits some uncertainty. 
Most of the predictions located at the corner are associated with \textsc{Before}, but there also are several predictions scattered around it.
We then simplified the sentence structure by removing ``\emph{but four}'', 
and fed the modified sentence to the same model.
This time, the model predicted the right temporal relation with much lower uncertainty (Figure~\ref{fig:uncertain}(b)).

In another case study (Figure~\ref{fig:uncertain}(c)), the true label is ``\emph{depart}'' \textsc{after} ``\emph{reveal}''.
This test case is rather straightforward, because of the explicit temporal word ``before''.
The model predicted \textsc{After} with high confidence, as shown by the predictive probabilities cluster at the top of the simplex.
To show the impact of the temporal description, we swapped it from ``\emph{before}'' to ``\emph{after}'' and fed it to the same model.
The model recognized the reversed meaning and correctly predicted \textsc{Before} with low uncertainty (Figure~\ref{fig:uncertain}(d)).
The above cases demonstrate that the proposed model reacts to different inputs with reasonable uncertainty, on both the total and model uncertainty scores.

\section{Conclusion}

We propose a principled approach to incorporate knowledge for event temporal relation extraction named Bayesian-Trans, which models the relation representations in the MuRE translational model, as latent variables.
The latent variables are inferred through variational inference, during which commonsense knowledge is incorporated in the form of the prior distribution.
The experiments on MATRES, TCR, and TBD show that Bayesian-Trans achieves state-of-the-art performance.
Comprehensive analyses of the experimental results also demonstrate the characteristics and benefits of the proposed model.

\section*{Limitations}
Our approach takes an event pair as input for the prediction of their temporal relation. We observe that if two events reside in different sentences, the error rate increases by 19\%. A promising future direction is to construct a global event graph where temporal relations of any two events are refined with the consideration of global consistency constraints, for example, no temporal relation loop is allowed in a set of events. Our current work only deals with even temporal relations, it could be extended to consider other event semantic relations such as causal, hierarchical or entailment relations. The event temporal knowledge in this paper is derived from ATOMIC which can possibly be extended to more sources. Bayesian learning could also be extended to life-long learning. But we need to explore approaches to address the problem of catastrophic forgetting.
We didn't exhaustively investigate all the translational models due to the large volume of work in that area.
There might be a translational model which can achieve better performance, but the core idea of the proposed framework stays the same.

\section*{Ethical Considerations}
The goal of the proposed method is to understand the temporal relation between events based on the descriptions in the given text.
What the method can achieve in the most optimistic scenario is no more than giving the same text to a human reader and letting him or her explain the event relations.
Therefore, the ethical concerns only come from the data collection.
In this paper, we only use publicly available datasets which have already been widely used in the research field.
As for potential application, as long as the user collects the training data legally, the proposed method does not have the potential to have a direct harmful impact.

\section*{Acknowledgements}

This work was supported in part by the UK Engineering and Physical Sciences Research Council (grant no. EP/T017112/1, EP/V048597/1, EP/X019063/1). YH is supported by a Turing AI Fellowship funded by the UK Research and Innovation (grant no. EP/V020579/1). This work was conducted on the UKRI/EPSRC HPC platform, Avon, hosted in the University of Warwick’s Scientific Computing Group. XT was partially supported by the Research Development Fund (RDF) 2022/23 (University of Warwick): `\textit{An Event-Centric Dialogue System for Second Language Learners}'.

\bibliography{anthology,custom}
\bibliographystyle{acl_natbib}

\clearpage

\appendix

\setcounter{table}{0}
\renewcommand{\thetable}{A\arabic{table}}

\section{Maximum Mean Discrepancy (MMD)}
\label{appendix:mdd}
The Maximum Mean Discrepancy (MMD) can be unbiasedly estimated using the following equation \cite{nan-etal-2019-topic}:

\vspace{-8pt}
{\small
\begin{equation}
\begin{split}
    \widehat{\textrm{MMD}} &=\frac{1}{N( N-1)}\sum\limits _{n\neq m} f_k\left( z_{n} \ ,z_{m} \ \right) \nonumber\\
    &+\frac{1}{N( N-1)}\sum _{n\neq m} f_k\left(\tilde{z}_{n} ,\tilde{z}_{m}\right) 
    -\frac{2}{N^{2}}\sum _{n,m} f_k\left( z_{n} ,\tilde{z}_{m}\right)\nonumber
    \label{mmd equation}
\end{split}
\end{equation}}
\vspace{-12pt}

\noindent where $z_1,...,z_n$ are sampled from variational distribution $q_{\phi}$ and $\tilde{z}_1,...,\tilde{z}_m$ are sampled from prior distribution $p$, $f_k(\cdot)$ is inverse multiquadratic knernel $f_k(x,y)=\frac{C}{C+||x-y||^2_2}$ which is often chosen for high-dimensional Gaussians.

\begin{table}[h]
  \begin{center}
  \resizebox{0.75\columnwidth}{!}{
  \begin{tabular}{lccc}
  \toprule
  \bf Model & \bf Precision & \bf Recall & \bf F\textsubscript{1} \\ 
  \midrule 
  TransE & $\bm{80.0}$ & $84.1$ & $82.0$ \\
  MuRP & $75.4$ & $85.8$ & $80.2$ \\
  AttH & $76.2$ & $\bm{87.4}$ & $81.4$\\
  MuRE & $79.6$ & $86.0$ & $\bf{82.7}$ \\
  \bottomrule
  \end{tabular}}
  \end{center}
  \caption{The performance of the proposed Bayesian framework combined with different translational models on MATRES.} 
  \label{table:dif_trans}
\end{table}

\section{Hyperparameter Settings and Resource Consumption}
\label{appendix:hyperparameters}

We conducted a grid-search to determine the optimal hyperparameters and dimensionality of the relation embeddings.
The searching range for the dimension of the latent vector the transformation parameters is $[50, 200]$, with a step size of $50$.
As a result, on the MATRES, the dimension of the latent vector $z$ is $200$, the dimension of relation transformation vectors $t_r$ or matrices $W_r$ is $50$, the dropout rate is $0.1$.
On the TBD, the dimension of the latent vector $z$ is $100$, and the dimension of relation transformation vectors $t_r$ or matrices $W_r$ is $100$, dropout rate is $0$.
Based on the above settings, the number of parameters of the Bayesian-Trans is $443$ thousand (excluding the COMET-BART).
The COMET-BART encoder has $204$ million parameters.
The learning rate $\alpha_c$ for the context encoder is set to $1e-5$, while for other components of the architecture $\alpha=1e-3$.
To calibrate the influence of the regularization term, we set a scaling weight smoothly increasing from $1e-2$ to $2.0$ during training.
We ran the training for $60$ epochs which is enough for the model to converge, and evaluated on the validation set after each training epoch.

All the experiments were conducted on an Nvidia GeForce RTX 3090 GPU. On the TBD dataset, the average training time is $93$ seconds per epoch, while the inference time is $4$ seconds. On the MATRES dataset, the average training time is $74$ seconds per epoch, and the inference time is $7$ seconds.

\section{Comparison of Translational Models}
\label{sec:appendix_trans}

Table \ref{table:dif_trans} shows the performance on MATRES using different translational models in the Bayesian framework.
TransE \cite{Bordes2013TranslatingEF} is one of the most commonly used translational models, which only performs the addition transformation on the head event.
AttH \cite{chami-etal-2020-low} expands the idea of hyperbolic translational models by making the curvature learnable. It also introduces more types of transformation, reflection and rotation.
MuRE \cite{Balazevic2019MultirelationalPG} strikes a balance by conducting diagonal matrix transformation and addition transformation.
MuRP \cite{Balazevic2019MultirelationalPG} is the Poincar\`{e} version of MuRE, which projects the head and tail onto a Poincar\`{e} ball before performing scaling and addition. The score function of MuRP computes the Poincar\`{e} distance instead of the Euclidean distance.

We can observe that TransE performs well, beating the previous state-of-the-art ($F_1=81.7$) but gives slightly worse results compared to MuRE. Both translational models in the hyperbolic space, MuRP and AttH, are inferior to the Euclidean-based translational models. As MuRE gives more balanced precision and recall values, it is therefore adopted in our Bayesian learning framework.

\begin{table}[htb]
  \begin{center}
  \resizebox{0.8\columnwidth}{!}{
  \begin{tabular}{lrrr}
  \toprule
  \bf Relation & \bf Precision & \bf Recall & \bf F\textsubscript{1} \\ 
  \midrule
  \textsc{Before} & $82.66$ & $90.40$ & $86.35$ \\
  \textsc{After} & $75.24$ & $88.56$ & $81.36$ \\
  \textsc{Equal} & $0.00$ & $0.00$ & $0.00$ \\
  \textsc{Vague} & $33.33$ & $15.60$ & $21.25$\\
  \bottomrule
  \end{tabular}}
  \end{center}
  \caption{Performance on each relation type class on MATRES.} 
  \label{table:class-wise}
\end{table}

\begin{table}[htb]
  \begin{center}
  \resizebox{0.8\columnwidth}{!}{
  \begin{tabular}{lrrr}
  \toprule
  \bf Relation & \bf Precision & \bf Recall & \bf F\textsubscript{1} \\ 
  \midrule
  \textsc{Before} & $77.04$ & $62.98$ & $69.31$ \\
  \textsc{After} & $76.37$ & $62.20$ & $68.56$ \\
  \textsc{Simultaneous} & $33.33$ & $10.42$ & $15.87$ \\
  \textsc{Includes} & $36.84$ & $10.29$ & $16.09$ \\
  \textsc{IsIncluded} & $53.33$ & $10.39$ & $17.39$ \\
  \textsc{None} & $59.06$ & $83.80$ & $69.29$ \\
  \bottomrule
  \end{tabular}}
  \end{center}
  \caption{Performance on each relation type class on TimeBank-Dense.} 
  \label{table:class-wise_tbd}
\end{table}

\section{Class-Specific Results}
In Table \ref{table:class-wise} and \ref{table:class-wise_tbd}, 
We show the results obtained using Bayesian-Trans under each temporal relation class on MATRES and TimeBank-Dense, respectively. On MATRES, the performance on \textsc{Before} and \textsc{After} are significantly better than for the other two classes.
The model predicts no \textsc{Equal} labels, most likely caused by the scarce training data for this class.
Previous works in the literature \cite{han-etal-2019-joint} have also shown similar class-specific results, with models struggling the most on the prediction of \textsc{Equal} and \textsc{Vague} relations. Similar conclusions can be drawn from the TimeBank-Dense dataset, that Bayesian-Trans performs relatively well on the \textsc{Before},  \textsc{After} and \textsc{None} classes, but performs worse on the other three minority classes. 
\end{document}